\pdfoutput=1
\documentclass{article}
\usepackage[final,dblblindworkshop]{neurips_2026}
\usepackage[utf8]{inputenc}
\usepackage[T1]{fontenc}
\usepackage{amsmath,booktabs,microtype,array,tabularx,longtable,placeins}
\newcolumntype{Y}{>{\raggedright\arraybackslash}X}
\usepackage{natbib}
\usepackage{xcolor}
\usepackage{pgfplots}
\pgfplotsset{compat=1.18}
\usepackage[colorlinks=true,allcolors=blue]{hyperref}
\hypersetup{
  pdfauthor={Le Chen, Zishen Wan, Baixi Sun, Xiaolong Ma, Chih-Hsuan Yang, Feng Yan, Sheng Di, Franck Cappello, Rajeev Thakur},
  pdftitle={Measure Before You Manage: Evaluating Agent Working Memory in Coding Agents},
  pdfkeywords={Coding agents, Agent working memory, Context management, Empirical evaluation, Resource accounting}
}


\workshoptitle{AgenticOS: Co-designing Systems and ML Foundations of an OS Layer for Agentic AI}
\title{Measure Before You Manage:\\
Evaluating Agent Working Memory in Coding Agents}
\author{%
  Le Chen$^{1}$ \quad Zishen Wan$^{2}$ \quad Baixi Sun$^{1}$ \quad
  Xiaolong Ma$^{1}$ \quad Chih-Hsuan Yang$^{1}$ \\
  \bf Feng Yan$^{3}$ \quad Sheng Di$^{1}$ \quad Franck Cappello$^{1}$ \quad
  Rajeev Thakur$^{1}$ \\[4pt]
  \normalfont
  $^{1}$Argonne National Laboratory \quad
  $^{2}$Columbia University \quad
  $^{3}$University of Houston \\
}

\begin{document}
\maketitle
\begin{abstract}
Agent working memory is heterogeneous. Objects such as instructions, artifacts, tool outputs, and agent-generated state play different semantic roles and exhibit different size, retention, and representation profiles. Recent work has begun to explore memory-management mechanisms that account for such heterogeneity. This work focuses on semantic heterogeneity and studies how it should shape the management and evaluation of working memory in coding agents.
Across 55 archived coding-agent trajectories, we find that semantically different working-memory objects exhibit distinct retention and compression behavior. This heterogeneity motivates semantically informed memory management.
We study two semantically informed strategies: an object-aware compression policy and a retrieval-based policy. Their evaluation shows that calibration gains may not transfer to held-out tasks, and that equal token budgets do not imply equal delivered context or management cost. A real-system replay further exposes serving limits that nominal budgets alone do not capture.
Together, these results show why semantic structure matters for agent working memory and why evaluating memory-management strategies requires more than a nominal token budget. We organize these lessons into four levels: stored state, delivered context, management work, and task or process outcome.
\end{abstract}

\section{Introduction}
\label{sec:intro}

Agent working memory is heterogeneous. During a coding trajectory, instructions,
source artifacts, tool outputs, and agent-generated state play different semantic
roles and evolve differently over time. Their size, persistence, representation,
and lifecycle characteristics need not follow the same pattern. Yet
memory-management mechanisms commonly center on retention, compression,
eviction, or retrieval rules that may treat this heterogeneous state uniformly.
Whether and how semantic heterogeneity should shape the management of agent
working memory---and how such management should be evaluated---remains unclear.

We study these questions in coding agents. We first characterize working memory
at the level of typed objects, asking how different kinds of state differ in
size, retention, representation, and compression behavior. We then examine two
semantically informed management strategies: a calibration-informed
object-aware compression policy and a retrieval-based policy adapted from prior
work. Our goal is not to identify a universally winning policy, but to understand
what these strategies reveal about heterogeneous working memory and what
evidence is required to evaluate them reliably. Specifically, we ask:
\emph{(1) how does semantic heterogeneity manifest in coding-agent working
memory, and (2) what must be measured before attributing a benefit to a
memory-management strategy?}

Our contributions are threefold. First, we characterize semantic heterogeneity
in coding-agent working memory through typed-object accounting of size,
retention, representation, and compression behavior across 55 archived
trajectories. Second, we examine two semantically informed management
strategies---an object-aware compression policy and a retrieval-based
policy---as case studies of how heterogeneous working memory is managed in
practice. Their results illustrate how apparent gains may not transfer to
held-out tasks and how nominally equal budgets can still leave delivered context
and management work unequal. Third, we synthesize these observations into a
four-level evaluation framework that separates stored state, delivered context,
management work, and task or process outcome.

\paragraph{Relation to existing work.}
Memory hierarchies and retrieval are established agent design ideas. MemGPT
manages movement between memory tiers for document analysis and multi-session
conversation \citep{packer2023memgpt}; Generative Agents combines memory
retrieval with reflection and planning in a social simulation
\citep{park2023generative}; and SWE-agent shows that the agent--computer
interface affects software-engineering behavior and performance
\citep{yang2024sweagent}. These works motivate mechanisms relevant to our
setting, but are not the systems evaluated here.

Prior work has also studied the evaluation and cost of agent memory.
\citet{lindenbauer2025complexity} compare observation masking and LLM
summarization on SWE-bench Verified; \citet{chintalapati2026condensation}
compare memory condensers on scientific-discovery tasks, reporting quality and
token costs; and \citet{omri2026agent} characterize construction, retrieval,
and generation costs across agent-memory systems. We do not claim novelty in
memory hierarchies, retrieval, summarization, or agent-memory profiling itself.
Our focus is narrower: task-local working memory in coding agents and how its
semantic heterogeneity should be measured, managed, and evaluated. We combine
typed-object residency accounting with explicit controls for calibration,
delivered context, management work, and outcome interpretation within a single
coding-agent archive. The empirical scope and outcome validation remain
deliberately limited.

\section{Working-Memory Model and Study Design}
\label{sec:setup}
\paragraph{Objects and agent interface.}
The host maintains an ordered message plan referencing the typed working-memory
objects. Instructions hold protected system/task text; artifacts include
source-code views; tool outputs hold execution feedback; agent state holds
model-generated text, not verified reasoning. Records include object ID, size, creation step,
representation, and optional path/version/dependencies. Compression policies
use raw, compressed, summary, and pointer representations; a pointer retains an
ID recoverable through \texttt{recall\_object}, so eviction need not be
irreversible deletion.

The policy studies use a 25-step ceiling and recorded model alias
\texttt{claude-opus-4.8}, with the gateway prefix anonymized; the served
revision is unpinned. Agents receive issue text, repository state, and
read/edit/search/list/test/command/recall tools, not a gold patch. Historical
requests, retries, and outside human supervision are incompletely
reconstructable, so we claim neither a wholly unattended workflow nor identical
historical serving conditions; auxiliary requests omit temperature, which does
not establish deterministic decoding.

\begin{table}[!ht]
\centering\small
\caption{Study populations and comparison units. These samples must not be
added together as independent replications. The retrieval study reuses
development tasks rather than providing another held-out test.}
\label{tab:study_design}
\begin{tabularx}{\linewidth}{@{}l r Y@{}}
\toprule
Study & Tasks & Evidence and conditioning \\
\midrule
Object characterization & 55 & One archived full-context trajectory per task;
eight repositories; descriptive object accounting. \\
Policy calibration & 15 & SymPy; task-level averages over nominal-budget
conditions at most 15\%. \\
Policy held-out & 8 & Five other repositories; one run per task--arm in a
separate 15\%-labelled sweep. \\
Retrieval follow-up & 24 & SymPy development tasks; eight blocks completed
all six arms once, yielding 48 paired-analysis cells. \\
\bottomrule
\end{tabularx}
\end{table}

\paragraph{Populations and outcomes.}
Table~\ref{tab:study_design} summarizes the study populations and comparison
units. Tasks derive from SWE-bench Lite \citep{jimenez2024swebench}, but the earlier
policy population was selected for full-context solvability under a local,
Docker-free evaluator checking \texttt{FAIL\_TO\_PASS} and only the first 20
\texttt{PASS\_TO\_PASS} tests, so its verdicts are not official SWE-bench
scores; the later 48 complete task--arm cells lack valid formal repair
evaluation. Our primary endpoint is therefore explicitly a level-4 \emph{process
metric}: each tool invocation whose tool name and sorted-key JSON arguments
match a previous invocation contributes one repeated call. A test after an edit
can legitimately repeat, so this metric measures process regularity rather than
repair success or total unnecessary work; an agent finish signal does not fill
that gap.

\paragraph{Content accounting.}
We deduplicate object IDs within each characterization run and exclude
disk-only records. For object $o$, let $s_o$ be its recorded proxy-token size,
$c_o$ its creation step, $e_o$ its first eviction, and $T$ the trajectory
endpoint ($e_o=T$ when no eviction is recorded):
\begin{equation}
r_o=\max(0,\min(e_o,T)-c_o),\qquad
V_c=\sum_{o\in c}s_o,\qquad R_c=\sum_{o\in c}s_or_o .
\label{eq:accounting}
\end{equation}
Content volume $V_c$ counts each retained record once; retention-weighted cost
$R_c$ weights it by recorded residency. The earlier host uses
\texttt{cl100k\_base} proxy accounting. Both are accounting quantities: neither
is a provider bill, a KV-cache footprint, or a measure of causal utility, and
equal-content records with different IDs are not deduplicated
(Appendix~\ref{app:setup} gives task composition and archive boundaries);
serving-side quantities were measured separately (Section~\ref{sec:resources}).

\section{Semantic Heterogeneity in Agent Working Memory}
\label{sec:characterization}
\begin{figure}[!ht]
\centering
\usetikzlibrary{patterns}
\definecolor{volumeblue}{RGB}{68,119,170}
\definecolor{costorange}{RGB}{221,132,82}
\begin{tikzpicture}
\begin{axis}[
  width=.94\linewidth,height=3.6cm,
  ybar,bar width=15pt,
  ymin=0,ymax=65,ytick={0,20,40,60},
  ylabel={Share (\%)},
  symbolic x coords={Tool output,Artifact,Instruction,Agent state},
  xtick=data,enlarge x limits=.18,
  axis x line*=bottom,axis y line*=left,
  ymajorgrids=true,grid style={black!12},
  tick label style={font=\small},
  label style={font=\small},
  nodes near coords,
  nodes near coords style={font=\footnotesize},
  every node near coord/.append style={/pgf/number format/fixed,/pgf/number format/precision=1},
  legend image code/.code={\draw[#1] (0cm,-0.07cm) rectangle (0.28cm,0.07cm);},
  legend cell align=left,
  legend style={at={(.5,1.04)},anchor=south,legend columns=2,draw=none,font=\small},
]
\addplot[fill=volumeblue,draw=black!75] coordinates {
 (Tool output,55.5) (Artifact,28.3) (Instruction,9.1) (Agent state,7.1)
};
\addplot[fill=costorange,draw=black!75,
  postaction={pattern=north east lines,pattern color=black!45}] coordinates {
 (Tool output,40.2) (Artifact,38.9) (Instruction,12.9) (Agent state,7.9)
};
\legend{Content volume,Retention-weighted cost}
\end{axis}
\end{tikzpicture}
\caption{Pooled object accounting across 55 coding trajectories. Each series
has its own denominator: summed content volume or summed retention-weighted
cost. These are descriptive shares, not task-level confidence intervals.
The same four object types contribute different shares when residency is
included.}
\label{fig:volume_residency}
\end{figure}
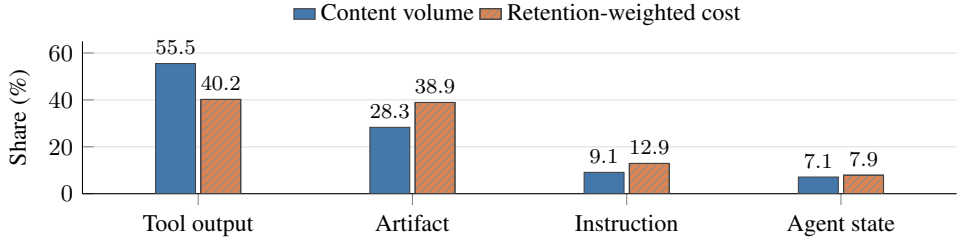

\paragraph{Object types differ in volume and residency.}
The 55 trajectories contain 1,350 in-context objects after excluding 306
disk-only records: 585 tool outputs, 165 artifacts, 110 instructions, and
490 agent-state objects. Tool outputs account for 55.5\% of pooled content
volume and 40.2\% of retention-weighted cost; artifacts account for 28.3\%
and 38.9\%, respectively (Figure~\ref{fig:volume_residency}). Artifacts'
median recorded size is 624 proxy tokens versus 73 for tool outputs, and their
mean recorded residency is 10.71 versus 8.61 steps. Artifacts thus nearly reach
tool outputs' retention-cost share without reversing the ordering, so a
volume-only profile understates their contribution once residency is taken into
account.

\paragraph{The pooled average is not a typical task.}
Weighting tasks equally gives a mean tool-output volume share of 50.8\% rather
than the pooled 55.5\%, with a task-level P10--P90 range of 23.0--78.1\%: these
are different estimands, not inconsistent results. An additional 18-trajectory
Terminal-Bench probe has a 64.7\% mean tool-output share and 4.5\% artifact
share, versus 50.8\% and 29.1\% in SWE tasks; it is a descriptive boundary
check, not a policy-transfer experiment. Temporal summaries also change
population: all 55 trajectories contribute at step 5 but only nine have an LLM
step at step 25, so their changing composition cannot establish a within-task
trend over the corpus.

\paragraph{Object types also differ in representation behavior.}
A separate structural probe covers 156 sampled objects, with 468
raw/compressed/pointer rows. The mean compressed/raw size ratio is 0.150 for
artifacts and 0.673 for tool outputs, so a single stored-token figure hides how
much state each representation carries. These ratios are properties of the
implemented type-specific rules and sampled objects rather than intrinsic
compressibility, and the probe lacks task IDs, so it supports neither
task-clustered inference nor semantic-preservation claims. The archived
literal-string retention metric assigns 1 to empty fact sets; restricting to
nonempty sets changes the instruction mean from 0.862 to 0.311
(Appendix~\ref{app:characterization}). No stored-state quantity identifies
which object is safe to discard.

\section{Managing Heterogeneous Working Memory}
\label{sec:policies}
This heterogeneity suggests two semantically informed responses: change object
representations by type, or select objects by retrieval score. We examine each
as a case study of how semantic structure can inform working-memory
management, while using their evaluation to identify where apparent policy
gains can be misleading. We do not treat either as a universally superior
policy, and we do not pool them because the studies differ in budget, host
mechanism, and population.

\subsection{Object-Aware Compression}
\label{sec:calibration}
The object-aware policy (OA) is a hand-specified heuristic, not a learned
utility predictor. Its score combines type/subtype weights, recorded age and
access count, stale/supersession penalties, and inverse square-root rendered
size. At most four rounds demote low-score candidates through raw, compressed,
summary, and pointer forms; instructions and newly created objects are
protected. Calibration informed the recorded defaults, but a complete
parameter-search ledger is unavailable (Appendix~\ref{app:calibration} gives
the equation and coefficients).

FIFO evicts by creation time; the earlier LRU uses the host's inherited access
clock. Uniform compression (UC) applies an age threshold of three steps and
then FIFO eviction, sharing the same type-specific structural renderers: it is
the decision rule, not the codec, that is uniform. One instrumentation limit
matters throughout: automatic prompt inclusion updates the inherited access
clock, including for pointers, so LRU can collapse toward creation order and
OA's age/reuse inputs are not clean demand signals. Version-derived penalties
have a further limitation (Section~\ref{sec:resources}).

\begin{table}[!ht]
\centering\small
\caption{Task-paired repeated-call contrasts: OA--baseline for calibration and
held-out; GA--baseline for retrieval. Negative $\Delta$ favors OA or GA
respectively; $n/m$: tasks/nonzero differences. CIs are post-hoc task
bootstraps; exact two-sided signed-rank tests omit zero differences; Holm
correction is separate for each of the three comparison families. The rows are
not a cross-study ranking.}
\label{tab:policy_effects}
\begin{tabular}{@{}llrrlrr@{}}
\toprule
Study & Baseline & $n/m$ & Mean $\Delta$ & 95\% CI & Exact $p$ & Holm $p$ \\
\midrule
Calibration & FIFO & 15/12 & -1.633 & [-2.667, -0.700] & 0.0049 & 0.0146 \\
Calibration & LRU & 15/10 & -1.533 & [-2.633, -0.533] & 0.0215 & 0.0312 \\
Calibration & UC & 15/8 & -0.733 & [-1.233, -0.267] & 0.0156 & 0.0312 \\
Held-out & FIFO & 8/2 & -0.500 & [-1.250, +0.000] & 0.5000 & 0.5000 \\
Held-out & LRU & 8/4 & -1.875 & [-4.125, +0.000] & 0.2500 & 0.5000 \\
Held-out & UC & 8/5 & -1.000 & [-1.750, -0.375] & 0.0625 & 0.1875 \\
Retrieval & FIFO & 8/5 & -0.375 & [-1.250, +0.375] & 0.7500 & 1.0000 \\
Retrieval & LRU-D & 8/7 & -0.125 & [-2.125, +1.000] & 0.3594 & 1.0000 \\
Retrieval & UC & 8/6 & +0.125 & [-0.625, +0.750] & 1.0000 & 1.0000 \\
Retrieval & OA & 8/6 & +0.500 & [+0.000, +0.875] & 0.2188 & 0.8750 \\
 
\bottomrule
\end{tabular}
\end{table}

For the original nominal-budget analysis, runs at fractions at most 15\% are
averaged within task before pairing OA with each baseline. Six calibration
tasks contribute two budget-labelled runs and nine contribute one; held-out has
eight tasks with one run per arm. We enumerate sign assignments with midranks
for tied absolute differences and drop zero differences; Holm correction covers
three baselines within each split. Post-hoc 95\% CIs use 10,000 task bootstrap
samples and seed 20260828, describing task variation rather than
repeated-generation variability; repositories can induce further dependence.

OA's calibration contrast with FIFO is $-1.633$ repeated calls, whereas its
held-out contrast is $-0.500$ with only
two nonzero pairs (Table~\ref{tab:policy_effects}), and no held-out contrast
survives Holm correction. The held-out split changes the evidentiary status of
the calibration result: a policy tuned on development tasks can show a sizable
development contrast on the repeated-call metric that its own held-out sweep
does not confirm. A non-significant held-out result is not equivalence, and the
pattern neither proves that calibration caused overfitting nor shows that OA
matches a baseline. A held-out OA--UC bootstrap interval excludes zero while
its exact test does not reject; the interval is not an inversion of that test,
and sparse discrete outcomes make the distinction matter.

Budget floors further qualify the nominal analysis. A post-hoc calibration
subset restricted to actual budget/reference-peak ratios at most 30\% contains
eight tasks; all three Holm-adjusted $p$ values equal 0.09375. This changes
both the sample and the conditioning rule, so it neither replaces the original
analysis nor identifies budget mismatch as the cause of unconfirmed transfer.

\subsection{Retrieval-Based Memory Management}
\label{sec:retrieval}
The later study adapts Generative Agents' recency/relevance/importance
components \citep{park2023generative}, not its reflection, planning, or
social environment. The score is
\begin{equation}
S_o^{\mathrm{GA}}=0.5\,\widetilde{0.99^{\,s-\ell_o}}
  +3.0\,\widetilde{\cos(e_o,e_q)}+2.0\,\widetilde{I_o},
\label{eq:retrieval}
\end{equation}
where each component is min--max normalized over candidates, with constants
mapped to 0.5, and the clock $\ell_o$ updates on admission, not automatic
prompt inclusion. Local \texttt{BAAI/bge-small-en-v1.5} embeddings compare
candidate content with issue text and latest agent state, and importance uses a
separate model request for a 1--10 rating, cached within a run. These
coefficients, step-based decay, and coding inputs define an adaptation, not a
replication.

The six arms are an uncapped full-context reference, FIFO, LRU-demand (LRU-D),
adapted retrieval (GA), UC, and OA. FIFO, LRU-D, and GA share a raw/pointer
packer and may restore old pointer objects; LRU-D observes READ/RETRIEVE/UPDATE
demand events instead of automatic prompt inclusion. UC and OA retain inherited
compression ladders under a common-budget adapter that rejects
non-decreasing-cost transitions; their candidate sets exclude currently evicted
objects. The design holds the cap fixed while varying priority score,
representation, and restoration mechanism together---a deployment-style
comparison, not a score ablation. Of 24 task blocks, eight completed all six
arms, 13 stopped under a greedy floor rule, and three stopped after OA
exhausted its allowed transitions; of 77 started trajectories, 61 completed and
16 were interrupted, and only 48 belong to the paired six-arm blocks. The floor is a state-dependent construction, not a proven
global minimum, and its recorded excesses range from 1 to 78 tokens. Stops are
neither evidence of task infeasibility nor formal repair failures; later unrun
arms are not assigned zero outcomes.

Across the eight complete blocks, GA's mean repeated-call difference from FIFO
is $-0.375$ (95\% CI $[-1.250,0.375]$); none of four baseline contrasts
survives Holm correction. Each task--arm runs once, and an LRU-D count of seven
on one task and OA's only nonzero count of two on another show how few cases
drive arm means. The analysis is conditional on completion and does not
establish a stable ranking over all 24 tasks (Appendix~\ref{app:retrieval}).

\section{Lessons for Evaluating Agent Working Memory}
\label{sec:resources}
Both studies were run under nominal token budgets and summarized by process
metrics. Examining what those comparisons actually controlled yields the
following lessons for evaluating working-memory management.

\paragraph{Nominal budget labels need not reflect effective budgets.}
A budget label is a three-part quantity---nominal fraction, absolute cap, and
measurement unit---whose parts move independently. Earlier sweeps use
$B=\max(\lfloor fP\rfloor,B_{\min})$, with reference peak $P$ and a minimum
budget; the wide sweep's 6,000-token floor makes all 25\%/50\% pairs identical
absolute-budget conditions, and subsequent sweeps in that study use a
1,200-token floor, so such labels are sweep coordinates rather than distinct
pressure levels or independent tasks. The retrieval study freezes
$B_t=\max(\lfloor0.15P_t\rfloor,\lfloor F_t/0.8\rfloor+1)$ from archived
reference-peak and floor statistics, and 18 of 24 caps exceed the nominal 15\%
anchor. All five constrained arms share this task-specific cap, measured by
tokenizing the joined rendered message plan with a frozen Qwen tokenizer; that
unit differs from earlier proxy accounting and excludes some full-request
overhead, and API prompt usage is recorded separately. Conflicting archived
narrative and cell-derived medians remain distinguished
(Appendix~\ref{app:resources}).

\paragraph{Shared caps control admissible state but not delivered context.}
For each task--arm, take the median managed-state token count across steps, and
call a pair matched if the larger median is at most 1.10 times the smaller.
None of the ten constrained-arm pairs matches on all eight complete tasks;
FIFO--GA matches on six of eight, with a largest gap of 18.7\%. On these eight
tasks, with one run per arm, a contrast at a common cap is therefore partly a
contrast in delivered context.

The distinction also appears at the serving boundary. In an exploratory replay
on an NVIDIA GB10 serving Qwen2.5-Coder-32B under a frozen 32,768-token limit,
the unconstrained arm reached 37,883 tokens on one task, so 6 of its 25 steps
exceeded that limit, while every constrained arm stayed at or below 16,643
tokens: a hard feasibility boundary on delivered context, not a hardware-memory
advantage for any policy (Appendix~\ref{app:hardware}).

\begin{table}[!ht]
\centering\small
\caption{Measured resources in the eight complete retrieval-study blocks. Main
calls count recorded agent LLM steps; input is summed main-agent API prompt
usage, excluding auxiliary requests; wall time is median whole-run seconds;
ceiling counts runs reaching 25 steps, not successful repairs.}
\label{tab:main_resources}
\begin{tabular}{@{}lrrrrrr@{}}
\toprule
Arm & Main calls & Importance & Summary & Input tokens & Wall (s) & Ceiling \\
\midrule
Full & 181 & 0 & 0 & 2,140,636 & 131.7 & 6/8 \\
FIFO & 191 & 0 & 0 & 1,147,832 & 103.0 & 7/8 \\
LRU-D & 181 & 0 & 0 & 1,074,649 & 99.5 & 4/8 \\
GA & 198 & 285 & 0 & 1,206,642 & 143.9 & 7/8 \\
UC & 197 & 0 & 0 & 1,091,350 & 102.0 & 7/8 \\
OA & 190 & 0 & 169 & 1,166,383 & 154.4 & 7/8 \\
 
\bottomrule
\end{tabular}
\end{table}

\paragraph{Management work is part of the comparison.}
GA adds 285 importance calls and OA 169 summary calls
(Table~\ref{tab:main_resources}); CPU embedding work is additional and
uncounted. GA--FIFO whole-run wall-time differences are positive on all eight
tasks (mean $+67.45$ seconds), though they include changed tool paths and
cannot isolate rating latency. The archive's two full-run dollar estimates,
\$213.52 and \$235.99, conflict and use preset prices, so neither is a billed
measurement nor a cost total for this table.

\paragraph{Lifecycle instrumentation must preserve the meaning of signals.}
In one saved SymPy trajectory, reading unchanged source creates a new disk
artifact version and triggers five tool-output invalidations, while equal
full-source hashes establish that the content did not change; a version event
can therefore be a rendering event rather than semantic staleness, although one
case does not estimate prevalence. Separately, prompt inclusion refreshes the
inherited access clock, conflating rendering with demand. Both can affect OA;
later LRU-D changes the clock mechanism but is not a corrected-OA experiment,
and the archive holds no corrected-system trajectories
(Appendix~\ref{app:validity}).

\section{A Framework for Evaluating Agent Working Memory}
\label{sec:lessons}
Table~\ref{tab:framework} synthesizes these lessons into what to report at each
level when evaluating agent working memory; it is development-informed,
synthesized from the case studies rather than specified in advance. Two
qualifications do not fit it:
calibration is development evidence, so a development contrast needs its own
held-out assessment; and mechanism defects such as the access-clock and version
signals qualify a comparison rather than explain it.

\begin{table}[!ht]
\centering\footnotesize
\caption{The four reporting levels, synthesized post-hoc from
Sections~\ref{sec:characterization}--\ref{sec:resources}.}
\label{tab:framework}
\begin{tabularx}{\linewidth}{@{}>{\raggedright\arraybackslash}p{0.16\linewidth} Y Y@{}}
\toprule
Level & Report & Reading it guards against \\
\midrule
Stored state & Type, size, representation, residency; pooled and task-level
shares & Volume as the whole stored-state profile \\
Delivered context & Nominal fraction, absolute cap, unit, delivered state per
arm & A shared cap as matched delivered context \\
Management work & Auxiliary calls by type, embeddings, wall time & A same-cap
comparison as same compute \\
Task/process outcome & Process metrics with valid task outcomes; stopping
outcomes, paired units, uncertainty & Repeated calls or finish signals as
repair success \\
\bottomrule
\end{tabularx}
\end{table}

\paragraph{Scope and limitations.}
The evidence base is small, repository-clustered, conditional on completion,
and limited to eight complete six-arm retrieval blocks with one run per
task--arm; the later study has no valid repair-success evaluation. The served
model revision is unpinned, the request and intervention history is
incompletely reconstructable, the OA lifecycle signals are defective, and no
corrected-OA trajectories exist. Policy repairs, cache benefits, serving
speedups, and semantic-utility claims are outside the evidence, and
Appendix~\ref{app:validity} states what the archive does and does not
reproduce.

\paragraph{Conclusion.}
Coding-agent working memory is not a uniform token pool: semantically
different objects exhibit distinct retention and representation behavior,
motivating management mechanisms that account for this structure. Our two case
studies also show that apparent policy gains depend on what is actually
delivered and what management work is required. Evaluating such strategies
therefore requires four measurements: stored state, delivered context,
management work, and task or process outcome. Measure before you manage.
\label{body:end}

\clearpage
\bibliographystyle{plainnat}
\bibliography{references_regular}
\clearpage
\appendix
\numberwithin{equation}{section}
\renewcommand{\thetable}{\Alph{section}.\arabic{table}}
\renewcommand{\theHtable}{\Alph{section}.\arabic{table}}
\section{Experimental Setup and State Interface}
\label{app:setup}
\setcounter{table}{0}

All expanded results use saved records, not new agent runs. Independent
recalculations, new descriptive summaries, and confidence intervals are
post-hoc audit analyses, not additional confirmatory experiments. Historical
protocol declarations and later amendments are distinguished below.

\subsection{Populations and evaluation units}
\begin{table}[!ht]
\centering\small
\caption{Study populations are not interchangeable. A task, a trajectory, an
object, and a task--policy cell are different units.}
\label{tab:populations}
\begin{tabularx}{\linewidth}{@{}l r Y@{}}
\toprule
Study & Tasks & Population and use \\
\midrule
SWE characterization & 55 & One archived full-context trajectory per task;
eight repositories; object accounting. \\
Terminal-Bench probe & 18 & Separate archived task trajectories; descriptive
composition comparison only, not a policy replication. \\
Policy calibration & 15 & SymPy; six tasks in the tight-budget sweep plus
nine further calibration tasks; nominal budgets at most 15\%. \\
Policy held-out & 8 & Five repositories; a separate 15\%-labelled sweep;
conditional policy comparison. \\
Retrieval follow-up & 24 & SymPy development set; eight task blocks completed
all six arms once, yielding 48 cells for paired comparisons. \\
\bottomrule
\end{tabularx}
\end{table}

Within the populations of Table~\ref{tab:populations}, the 55-trajectory SWE
corpus contains SymPy (31), pytest (8), seaborn (4),
pylint (4), Sphinx (4), Django (2), Flask (1), and requests (1).
The collection summaries cover 24 calibration and 30 later tasks; the extra
trace \texttt{sympy\_\_sympy-11400} has no corresponding summary verdict.
It contributes to cost accounting, not to a claimed success rate. The eight
held-out policy tasks come from seaborn (2), pylint (2), pytest (2), requests
(1), and Sphinx (1). The retrieval follow-up uses development tasks and is
not a second held-out evaluation.

The earlier policy population was selected using successful full-context runs
under a local, Docker-free evaluator. Its base validation checks
\texttt{FAIL\_TO\_PASS} and only the first 20 \texttt{PASS\_TO\_PASS} tests.
These local verdicts are not presented as official SWE-bench scores
\citep{jimenez2024swebench}. Runs at different budgets are aggregated within
task before paired inference. The follow-up's eight-task estimand is additionally
conditional on all six arms completing under the stopping protocol.

\subsection{Object and agent interfaces}
The ordered message plan refers to object IDs with type/subtype, creation step,
size, representation, and optional path/version/dependencies. Instructions
hold protected system/task text; artifacts include code views; tool outputs
hold execution feedback; agent state holds model-generated text, not
validated reasoning. Disk-only source versions are excluded from context
cost. A pointer retains an ID that \texttt{recall\_object} can restore;
eviction is not permanent deletion.

Policy trajectories use a 25-step ceiling and the recorded model alias
\texttt{claude-opus-4.8}; its gateway prefix is withheld for double-blind review.
The agent receives issue text, repository state,
and read/edit/search/list/test/command/recall tools, not a supplied gold patch.
The alias does not pin the served revision. Auxiliary requests omit
temperature, which is not deterministic decoding. Outside human supervision
is not fully reconstructable; no wholly unattended workflow is claimed.

\paragraph{Primary process endpoint.}
Each tool invocation matching a previous tool name and sorted-key JSON
argument serialization contributes one repeated call. Changed arguments form
a new signature; a useful retest after an edit can still count as repeated.
Neither this endpoint nor a finish signal establishes repair success.

\FloatBarrier
\section{Extended State Characterization}
\label{app:characterization}
\setcounter{table}{0}

\subsection{Size, retention, and task heterogeneity}
For each trajectory, object records are deduplicated by ID, not by equal
content. The analysis keeps \texttt{in\_context=true}, giving 1,350 objects
and excluding 306 disk-only records. An object's proxy size uses the host's
\texttt{cl100k\_base} accounting. Its recorded retention interval is
$r_o=\max(0,\min\{e_o,T\}-c_o)$, where $c_o$ is creation, $T$ is the
trajectory endpoint, and $e_o$ is first eviction, or $T$ if absent.
Thus $s_or_o$ is an interval-based token-step proxy, not a sum of provider
charges and not the period for which the object is semantically useful;
Table~\ref{tab:size_details} reports the resulting size and retention
distributions by type.

\begin{table}[!ht]
\centering\small
\caption{Object-size distributions and recorded retention in the SWE corpus.
P90 is a linearly interpolated descriptive percentile; objects are not treated
as independent experimental replicates.}
\label{tab:size_details}
\begin{tabular}{@{}lrrrrr@{}}
\toprule
Type & Objects & Median size & P90 size & Mean $r_o$ & $\sum s_or_o$ \\
\midrule
Tool output & 585 & 73 & 931.6 & 8.61 & 1,599,232 \\
Artifact & 165 & 624 & 1596.8 & 10.71 & 1,547,083 \\
Instruction & 110 & 135 & 787.3 & 13.25 & 514,522 \\
Agent state & 490 & 37 & 162.7 & 8.74 & 313,525 \\
 
\bottomrule
\end{tabular}
\end{table}

Pooled shares in the main text divide summed content volume by total volume.
Table~\ref{tab:task_shares} instead gives every task equal weight. For example,
tool output contributes 55.5\% of pooled volume but 50.8\% of the mean
within-task share. These are different estimands, not conflicting measurements.
The distributions show variation that a single pooled percentage conceals.

\begin{table}[!ht]
\centering\small
\caption{Within-task content-volume shares (\%). SWE statistics use 55 tasks;
the last column uses 18 separate Terminal-Bench trajectories, including zero
shares when a type is absent. P10--P90 is a descriptive range, not a confidence interval.}
\label{tab:task_shares}
\begin{tabular}{@{}lrrrr@{}}
\toprule
Type & SWE mean & SWE median & SWE P10--P90 & Terminal mean \\
\midrule
Tool output & 50.8 & 48.8 & [23.0, 78.1] & 64.7 \\
Artifact & 29.1 & 28.3 & [11.0, 45.4] & 4.5 \\
Instruction & 14.6 & 8.5 & [2.4, 35.0] & 17.7 \\
Agent state & 5.5 & 3.2 & [1.3, 11.8] & 13.0 \\
 
\bottomrule
\end{tabular}
\end{table}

The Terminal-Bench probe has more tool-output volume and less artifact volume
than the SWE sample. It is a boundary observation: repository work and terminal
tasks need not induce the same mix. It does not establish transfer of either
policy or consistency of semantic-staleness behavior across workloads.

\subsection{Composition over the recorded trajectory}
\begin{table}[!ht]
\centering\small
\caption{Mean within-step attributed-context shares (\%) at selected
checkpoints, recomputed post hoc. Each row averages only trajectories with a
recorded LLM step at that checkpoint. The changing denominator is essential.}
\label{tab:temporal}
\begin{tabular}{@{}rrrrrr@{}}
\toprule
Step & Trajectories & Tool output & Artifact & Instruction & Agent state \\
\midrule
1 & 55 & 0.0 & 0.0 & 100.0 & 0.0 \\
5 & 55 & 18.2 & 40.1 & 27.8 & 14.0 \\
10 & 25 & 35.2 & 36.3 & 11.6 & 16.9 \\
15 & 19 & 26.4 & 40.7 & 8.6 & 24.3 \\
20 & 13 & 35.9 & 33.6 & 5.7 & 24.8 \\
25 & 9 & 31.5 & 36.0 & 4.9 & 27.6 \\
 
\bottomrule
\end{tabular}
\end{table}

Unlike the once-per-object volume measure, Table~\ref{tab:temporal} describes
the state included at individual LLM steps. Later rows contain fewer,
longer-running tasks; their shifts cannot be interpreted as a balanced
longitudinal effect over all 55 tasks. In particular, a rising agent-state
share among survivors does not demonstrate rising utility.

\subsection{Structural compression probe}
The archived probe contains 468 raw/compressed/pointer rows for 156 sampled
objects. No summary-model probe results are present in this file. The sampler
uses available content and per-type limits, rather than a random task sample;
it does not enforce the characterization table's in-context filter. Because
the CSV lacks task IDs, these rows cannot be assumed to be a mapped subset
of the 1,350 context objects or independent task-level replicates.

The structural rules retain code signatures/imports for artifacts, selected
diagnostic lines for tool outputs, and a head portion for instructions and
agent state. Different ratios therefore describe these type-dependent rules
on this sample, not an intrinsic or optimal compressibility of each type.
The literal-fact metric extracts strings such as paths, error names, symbols,
test names, and test-count phrases and checks whether they remain verbatim.

\begin{table}[!ht]
\centering\small
\caption{Descriptive structural compression. Ratio is compressed/raw proxy
tokens. ``All'' is the archived mean literal-string retention; empty fact
sets receive 1 by convention. ``Nonempty'' is a new post-hoc sensitivity
summary restricted to rows with at least one extracted fact.}
\label{tab:compression_probe}
\begin{tabular}{@{}lrrrrr@{}}
\toprule
Type & Objects & Ratio & All & Empty fact sets & Nonempty \\
\midrule
Tool output & 40 & 0.673 & 0.660 & 7 & 0.588 \\
Artifact & 40 & 0.150 & 0.838 & 0 & 0.838 \\
Instruction & 40 & 0.518 & 0.862 & 32 & 0.311 \\
Agent state & 36 & 0.970 & 0.972 & 24 & 0.917 \\
 
\bottomrule
\end{tabular}
\end{table}

The 0.150 artifact ratio versus 0.673 for tool outputs in
Table~\ref{tab:compression_probe} supports considering
type-specific representations. It does not show semantic preservation or
unchanged task performance. Empty fact sets are common in instructions and
agent state, limiting the interpretation of their high ``All'' values.
We omit the old object-level significance test: task clustering cannot be
reconstructed reliably from the stored probe table alone.

\FloatBarrier
\section{Calibration-Informed Object-Aware Policy}
\label{app:calibration}
\setcounter{table}{0}

\subsection{Implemented scoring and baselines}
The policy is a hand-specified heuristic, not a learned estimator of future
utility. For type $t$, subtype multiplier $m_o$, recorded access age $a_o$,
access counter $k_o$, and recorded stale indicator $z_o$, the implementation uses
\begin{align}
u_o &= \max\!\left(10^{-6},
b_t e^{-d_ta_o}m_o[1+0.35\min(k_o,5)](1-p_tz_o)q_o\right),\\
S_o &= u_o/\max(s_o^{\mathrm{rendered}},1)^{0.5}.
\end{align}
Here $q_o=0.25$ for an artifact whose version is older than the store's current
version for that path, and 1 otherwise. Table~\ref{tab:oa_profile} gives the
recorded defaults. Subtype multipliers are diagnosis 1.4, plan 1.2, test 1.0,
compiler 0.9, search 0.7, listing 0.4, file\_read 0.9, and patch 1.1; unlisted
subtypes use 1.0. These are calibrated design choices, not measured causal values.

\begin{table}[!ht]
\centering\small
\caption{Recorded type profile. Instructions are pinned, so their profile is
not an instruction-eviction experiment.}
\label{tab:oa_profile}
\begin{tabular}{@{}lrrr@{}}
\toprule
Type & Base $b_t$ & Decay $d_t$ & Stale penalty $p_t$ \\
\midrule
Instruction & 1.0 & 0.00 & 0.0 \\
Agent state & 0.8 & 0.04 & 0.2 \\
Artifact & 0.6 & 0.06 & 0.7 \\
Tool output & 0.5 & 0.18 & 0.9 \\
\bottomrule
\end{tabular}
\end{table}

At most four rounds demote low-score candidates one rung through raw,
compressed, summary, and pointer until the budget is met. Candidates exclude
instructions and objects created in the current step. FIFO evicts by creation
order; the earlier LRU uses the inherited access clock. Uniform compression
(UC) applies a common age threshold of three steps and then FIFO eviction,
but shares the type-dependent structural renderers; ``uniform'' refers to the
decision rule, not one type-blind codec.

The earlier host increments access clocks and counters on automatic prompt
inclusion. Consequently, these are not clean demand-reuse signals: inherited
LRU can collapse to creation order, and OA's access-age and reuse terms are
degenerate under repeated inclusion. The version issue in
Appendix~\ref{app:validity} also affects its stale/supersession inputs. The
tables report the implemented historical policies, not their intended or
corrected variants. The archive documents fixed defaults and development use,
but not a complete hyperparameter-search ledger; we do not claim exhaustive tuning.

\subsection{Paired primary comparisons}
For the original nominal-budget analysis, each task's repeated-call counts
are first averaged across its cells with nominal fraction at most 15\%.
OA--baseline differences are then paired by task. The calibration set has
15 tasks; six contribute two budget-labelled runs and nine one. Held-out
uses one run per arm for each of eight tasks. Repeated budget conditions are
not treated as independent tasks.

\begin{table}[!ht]
\centering\small
\caption{OA minus each baseline on repeated tool calls. Negative favors OA.
$n/m$ gives paired tasks/nonzero differences. CIs are independently recomputed
post-hoc task-bootstrap intervals, not an inversion of the signed-rank test.}
\label{tab:h6_full}
\begin{tabular}{@{}llrrlrr@{}}
\toprule
Split & Baseline & $n/m$ & Mean $\Delta$ & 95\% CI & Exact $p$ & Holm $p$ \\
\midrule
Calibration & FIFO & 15/12 & -1.633 & [-2.667, -0.700] & 0.0049 & 0.0146 \\
Calibration & LRU & 15/10 & -1.533 & [-2.633, -0.533] & 0.0215 & 0.0312 \\
Calibration & UC & 15/8 & -0.733 & [-1.233, -0.267] & 0.0156 & 0.0312 \\
Held-out & FIFO & 8/2 & -0.500 & [-1.250, +0.000] & 0.5000 & 0.5000 \\
Held-out & LRU & 8/4 & -1.875 & [-4.125, +0.000] & 0.2500 & 0.5000 \\
Held-out & UC & 8/5 & -1.000 & [-1.750, -0.375] & 0.0625 & 0.1875 \\
 
\bottomrule
\end{tabular}
\end{table}

The two-sided exact test behind Table~\ref{tab:h6_full} drops zero differences,
assigns midranks to tied
absolute differences, and enumerates all sign assignments. Holm correction
covers the three baseline comparisons separately within each split. It was
an additional multiplicity analysis, not part of the original preregistration.
Intervals use 10,000 task bootstrap samples, seed 20260828, and sorted
bootstrap-mean indices 250 and 9750. Discreteness and the different procedures
can yield a bootstrap interval excluding zero without a significant exact
test, as in held-out OA--UC. Neither procedure establishes equivalence.

\subsection{Per-task results and secondary measurements}
\begin{table}[!ht]
\centering\small
\caption{Every task in the nominal primary comparisons: within-task mean
repeated calls. Calibration IDs have prefix \texttt{sympy\_\_sympy-}; held-out
IDs show repository and issue suffix. Fractional values average two
budget-labelled runs. UC = uniform compression; OA = object-aware.}
\label{tab:h6_tasks}
\begin{minipage}[t]{0.43\linewidth}
\centering
\emph{Calibration ($n=15$)}\par\smallskip
\begin{tabular}{@{}lrrrr@{}}
\toprule
Task & FIFO & LRU & UC & OA \\
\midrule
12481 & 2 & 1.5 & 0.5 & 1 \\
13647 & 0 & 3 & 0 & 0 \\
13773 & 1 & 0 & 1 & 0 \\
14817 & 2 & 2 & 1 & 0 \\
15609 & 1 & 0 & 0 & 0 \\
17022 & 4 & 4 & 2 & 0.5 \\
17139 & 0 & 0 & 1 & 1 \\
18189 & 0 & 3 & 2 & 0 \\
19487 & 2 & 0 & 0 & 0 \\
20154 & 5.5 & 5.5 & 3 & 0 \\
20442 & 6 & 6 & 2 & 0 \\
21055 & 3.5 & 2.5 & 2.5 & 1.5 \\
23262 & 0 & 0 & 0.5 & 0.5 \\
24152 & 0 & 0 & 0 & 0 \\
24213 & 2 & 0 & 0 & 0 \\
 
\bottomrule
\end{tabular}
\end{minipage}\hfill
\begin{minipage}[t]{0.55\linewidth}
\centering
\emph{Held-out ($n=8$)}\par\smallskip
\begin{tabular}{@{}lrrrr@{}}
\toprule
Task & FIFO & LRU & UC & OA \\
\midrule
seaborn-3010 & 2 & 1 & 2 & 2 \\
seaborn-3190 & 0 & 0 & 1 & 0 \\
requests-3362 & 0 & 8 & 1 & 0 \\
pylint-5859 & 1 & 0 & 2 & 0 \\
pylint-7228 & 0 & 0 & 1 & 0 \\
pytest-11148 & 0 & 5 & 0 & 0 \\
pytest-9359 & 5 & 5 & 5 & 2 \\
sphinx-10325 & 0 & 0 & 0 & 0 \\
 
\bottomrule
\end{tabular}
\end{minipage}
\end{table}

\begin{table}[!ht]
\centering\small
\caption{Descriptive secondary measurements under the same within-task
aggregation. Input is recorded main-agent prompt tokens per trajectory, not
an all-service token bill. No secondary significance claim is made here.}
\label{tab:h6_secondary}
\begin{tabular}{@{}llrrrr@{}}
\toprule
Split & Policy & Repeated & Tool calls & Steps & Input tokens \\
\midrule
Calibration & FIFO & 1.933 & 18.43 & 17.93 & 71,335 \\
Calibration & LRU & 1.833 & 19.70 & 19.37 & 78,917 \\
Calibration & UC & 1.033 & 19.47 & 18.30 & 74,612 \\
Calibration & OA & 0.300 & 17.67 & 17.30 & 72,422 \\
Held-out & FIFO & 1.000 & 20.12 & 19.88 & 99,480 \\
Held-out & LRU & 2.375 & 19.00 & 18.62 & 98,189 \\
Held-out & UC & 1.500 & 21.50 & 21.12 & 114,850 \\
Held-out & OA & 0.500 & 17.50 & 17.38 & 91,898 \\
 
\bottomrule
\end{tabular}
\end{table}

Calibration improvements on repeated calls do not imply improvement on all
resources: in Table~\ref{tab:h6_secondary}, OA's mean recorded input is
slightly above FIFO's under this aggregation. Held-out repeated-call
differences in Table~\ref{tab:h6_tasks} are driven by few nonzero
pairs, and neither fewer calls nor fewer steps demonstrates a repaired issue.
The local evaluator's limited scope and the later study's absent formal repair
evaluation prevent a cross-study success-rate claim.

\subsection{Post-hoc effective-budget sensitivity}
For each cell, effective fraction is its recorded absolute budget divided by
the full-context reference peak \emph{from the same sweep and task}.
The archived calibration sensitivity uses effective fraction at most 30\%,
not the primary nominal threshold of 15\%. Recomputing that archived selection
gives eight tasks and the comparisons in Table~\ref{tab:h6_effective}. Thus it
changes both the
conditioning rule and sample; it cannot replace the nominal primary analysis
or identify budget mismatch as the cause of failed held-out transfer.

\begin{table}[!ht]
\centering\small
\caption{Calibration sensitivity, post hoc, effective fraction $\leq30\%$.
All three Holm-adjusted values equal 0.09375. CIs use the same audit bootstrap
as Table~\ref{tab:h6_full}.}
\label{tab:h6_effective}
\begin{tabular}{@{}lrrlrr@{}}
\toprule
Baseline & $n/m$ & OA--baseline & 95\% CI & Exact $p$ & Holm $p$ \\
\midrule
FIFO & 8/7 & -2.312 & [-3.875, -0.812] & 0.0312 & 0.0938 \\
LRU & 8/7 & -2.500 & [-4.062, -0.938] & 0.0469 & 0.0938 \\
UC & 8/7 & -1.250 & [-2.000, -0.500] & 0.0312 & 0.0938 \\
 
\bottomrule
\end{tabular}
\end{table}

The separate, later 25\%/35\% held-out archive is not merged into these
eight-task results. Its different tasks, budgets, and full-context references
would define another analysis. These appendix computations do not open that
raw archive; pre-existing project summaries already contain some of its
derived results, so this is not a claim of prospective blinding of the write-up.

\FloatBarrier
\section{Retrieval-Based Adaptation and Six-Arm Evaluation}
\label{app:retrieval}
\setcounter{table}{0}

\subsection{What was adapted}
The follow-up adapts the recency/relevance/importance retrieval components of
Generative Agents \citep{park2023generative}, not its reflection, planning,
or simulated social environment. Our implemented score is
\begin{equation}
S_o^{\mathrm{GA}}=0.5\,\widetilde{0.99^{\,s-\ell_o}}
 +3.0\,\widetilde{\cos(e_o,e_q)}+2.0\,\widetilde{I_o}.
\end{equation}
Each tilde is min--max normalization over the current candidate set; a constant
component maps to 0.5. The retrieval-owned clock $\ell_o$ is initialized at
creation and updated for admitted objects, not on automatic prompt inclusion.
Elapsed agent steps substitute for the source setting's time scale.

Relevance uses local \texttt{BAAI/bge-small-en-v1.5}, revision
\texttt{5c38ec7c405e}\ldots, on CPU in float32 with normalized embeddings
and no query prefix. The query combines issue text and latest agent-state
text, with a nominal 512-token embedding-query allocation split between them
and unused allocation reassigned. Candidate embeddings use immutable content
and a content-hash cache. The importance request asks for one integer from
1 to 10 indicating future relevance to a software-engineering trajectory.
Its configuration is the recorded agent model alias, temperature omitted,
four output tokens, and one retry for an invalid rating. Long inputs retain
256 head and 256 tail Qwen tokens plus a separator; the delivered length is
logged rather than assumed to be exactly 512. Ratings are cached within a run.

\paragraph{Comparison mechanisms.}
FIFO, LRU-demand (LRU-D), and GA share a candidate universe and raw/pointer
packer. It includes old non-instruction objects even if currently pointers,
so retrieval may restore them. Instructions and newly created objects are
fixed. The packer starts with candidate pointers, makes a sequential
non-increasing-cost raw pass, and then offers candidates by priority, skipping
ones that do not fit. Rendering remains in message-plan order. FIFO prioritizes
newer creation; LRU-D uses READ/RETRIEVE/UPDATE demand events rather than prompt
inclusion. These are not the earlier FIFO/LRU implementations.

UC and OA keep the inherited compression mechanisms under a common budget
adapter: full-plan Qwen accounting and rejection of transitions that do not
strictly reduce that count. Their candidates exclude currently evicted objects;
their ladders and restoration opportunities differ from the selectors.
GA--compression comparisons therefore compare system mechanisms, not only a
retrieval score. Full context is an uncapped reference, not a sixth capped arm.

\subsection{Completed-block outcomes}
\begin{table}[!ht]
\centering\small
\caption{Repeated tool calls in all eight complete blocks. All task IDs have
prefix \texttt{sympy\_\_sympy-}; each task--arm combination ran once.
UC and OA here denote the common-budget-adapter variants.}
\label{tab:ga_matrix}
\begin{tabular}{@{}lrrrrrr@{}}
\toprule
Task & Full & FIFO & LRU-D & GA & UC & OA \\
\midrule
11870 & 0 & 1 & 0 & 1 & 0 & 2 \\
13146 & 0 & 0 & 0 & 1 & 1 & 0 \\
16106 & 0 & 0 & 0 & 1 & 0 & 0 \\
16281 & 0 & 3 & 7 & 0 & 2 & 0 \\
17022 & 0 & 1 & 0 & 1 & 0 & 0 \\
20154 & 0 & 1 & 0 & 0 & 1 & 0 \\
21627 & 0 & 2 & 0 & 1 & 0 & 0 \\
23262 & 0 & 1 & 0 & 1 & 1 & 0 \\
Mean & 0.000 & 1.125 & 0.875 & 0.750 & 0.625 & 0.250 \\
 
\bottomrule
\end{tabular}
\end{table}

\begin{table}[!ht]
\centering\small
\caption{All four declared GA--baseline contrasts ($n=8$). Negative favors GA;
$m$ counts nonzero differences. Exact tests and Holm values match the archived
analysis; intervals were independently recomputed with the audit bootstrap.}
\label{tab:ga_contrasts}
\begin{tabular}{@{}lrrlrr@{}}
\toprule
Baseline & $m$ & Mean $\Delta$ & 95\% CI & Exact $p$ & Holm $p$ \\
\midrule
FIFO & 5 & -0.375 & [-1.250, +0.375] & 0.7500 & 1.0000 \\
LRU-D & 7 & -0.125 & [-2.125, +1.000] & 0.3594 & 1.0000 \\
UC & 6 & +0.125 & [-0.625, +0.750] & 1.0000 & 1.0000 \\
OA & 6 & +0.500 & [+0.000, +0.875] & 0.2188 & 0.8750 \\
 
\bottomrule
\end{tabular}
\end{table}

In Table~\ref{tab:ga_matrix}, the largest single repeated-call count is seven
for LRU-D on task 16281;
OA's two calls on 11870 are its only nonzero entry.
Table~\ref{tab:ga_contrasts} reports the four declared GA--baseline contrasts;
the per-task table exposes
this concentration rather than interpreting the arm means as a stable ranking.
Full's zero is an observation, not a consequence of the metric definition.
Most arms often reach the step ceiling (Appendix~\ref{app:resources}).

\subsection{All attempted task blocks and stopping}
Table~\ref{tab:block_ledger} summarizes all 24 attempted blocks: eight
complete, 13 stopped by the greedy
floor rule, and three stopped after OA's adapted mechanism exhausted its
allowed transitions. No unresolved-infrastructure block is recorded. In total,
77 trajectories started, 61 completed, and 16 were interrupted. Only 48 of
those completed trajectories belong to complete six-arm blocks; the remaining
completed partial-block trajectories are not extra paired tasks. The protocol
stops a block after a failure, so later arms in that block can be unrun.

\begin{table}[!ht]
\centering\small
\caption{All 24 blocks, including non-completers. Task prefix is
\texttt{sympy\_\_sympy-}. State C = complete, F = floor rule, O = OA
exhaustion. $B/P$ uses the frozen reference peak. Excess is the recorded
greedy-floor count minus cap; a dash means not applicable, not zero.}
\label{tab:block_ledger}
\begin{minipage}[t]{0.48\linewidth}
\centering
\begin{tabular}{@{}rlrrr@{}}
\toprule
Task & State & $B$ & $100B/P$ & Excess \\
\midrule
11870 & C & 3037 & 21.7 & -- \\
12481 & O & 1887 & 15.0 & -- \\
13146 & C & 2709 & 19.8 & -- \\
13437 & O & 993 & 15.0 & -- \\
13647 & F & 1438 & 26.1 & 16 \\
13773 & F & 1156 & 19.2 & 9 \\
14817 & F & 1517 & 19.2 & 3 \\
15609 & F & 1002 & 15.0 & 1 \\
16106 & C & 2849 & 28.4 & -- \\
16281 & C & 3689 & 25.2 & -- \\
17022 & C & 3069 & 15.0 & -- \\
17139 & F & 1332 & 60.2 & 2 \\
 
\bottomrule
\end{tabular}
\end{minipage}\hfill
\begin{minipage}[t]{0.48\linewidth}
\centering
\begin{tabular}{@{}rlrrr@{}}
\toprule
Task & State & $B$ & $100B/P$ & Excess \\
\midrule
18057 & F & 1892 & 15.0 & 32 \\
18189 & F & 986 & 15.9 & 36 \\
18835 & F & 808 & 33.6 & 20 \\
19487 & F & 999 & 24.8 & 78 \\
20154 & C & 1682 & 17.2 & -- \\
20442 & F & 1034 & 24.2 & 68 \\
21055 & F & 1036 & 26.7 & 2 \\
21627 & C & 3494 & 15.0 & -- \\
22840 & O & 3458 & 44.3 & -- \\
23262 & C & 2172 & 36.6 & -- \\
24152 & F & 1222 & 76.7 & 12 \\
24213 & F & 926 & 39.4 & 12 \\
 
\bottomrule
\end{tabular}
\end{minipage}
\end{table}

``Floor rule'' refers to a sequential, state-dependent raw/pointer construction.
It is not a proven global minimum over representation assignments and does
not establish that the task is intrinsically infeasible. Likewise, OA
exhaustion concerns that implemented transition mechanism. The floor excess
ranges from 1 to 78 tokens; only three of the 13 cases are within two tokens.
The eight-completer analysis is conditional, not an estimate over all 24 tasks.
An interruption is not relabelled a repair failure, and an unrun arm is not
assigned a zero outcome.

\FloatBarrier
\section{Budget and Resource Accounting}
\label{app:resources}
\setcounter{table}{0}

\subsection{Nominal, absolute, and delivered quantities}
The earlier sweeps set $B=\max(\lfloor fP\rfloor,B_{\min})$, where $f$ is
the requested fraction and $P$ is the task's full-context reference peak in
that sweep. Table~\ref{tab:old_budgets} reconstructs $B/P$ from saved cells.
The original wide sweep used a 6,000-token floor; the tight and subsequent
15\% sweeps used 1,200. A nominal label is therefore not an actual pressure
level. For example, all 25\%/50\% pairs in the wide sweep received the same
absolute budget. They are repeated runs of the same budget condition, not
two pressure levels and not independent tasks.

\begin{table}[!ht]
\centering\small
\caption{Earlier sweeps: budget/reference peak (\%). ``Cells'' is the
median independently reconstructed from saved cells; ``Note'' is the
conflicting archived narrative median. $N$ counts constrained runs, not tasks.}
\label{tab:old_budgets}
\begin{tabular}{@{}lrrrrr@{}}
\toprule
Archive sweep & Nominal & $N$ & Cells & Note & Cell range \\
\midrule
policy & 25 & 24 & 71.1 & 73.5 & [56.3, 165.0] \\
policy & 50 & 24 & 71.1 & 73.5 & [56.3, 165.0] \\
policy\_tight & 8 & 24 & 15.6 & 16.3 & [11.9, 30.9] \\
policy\_tight & 15 & 24 & 15.6 & 16.3 & [15.0, 30.9] \\
policy\_w3 & 15 & 36 & 47.1 & 47.1 & [16.0, 63.1] \\
policy\_heldout & 15 & 32 & 25.7 & 32.7 & [15.0, 52.0] \\
 
\bottomrule
\end{tabular}
\end{table}

The older \path{FINDINGS_BUDGET_FLOOR.md} narrative and current saved cells
give different medians, as shown explicitly above; their historical discrepancy
is unresolved. Recalculations here use the saved cell values with within-sweep
references. No later 25\%/35\% held-out data are merged into this table.

The later study instead freezes, for each task $t$,
\begin{equation}
B_t=\max\!\left(\lfloor0.15P_t\rfloor,
                 \lfloor F_t/0.8\rfloor+1\right),
\end{equation}
where $P_t$ and $F_t$ are the archived reference peak and floor statistic used
by the freeze procedure. This is a budget-allocation rule, not a mathematical
infeasibility theorem. Eighteen of 24 tasks are raised above the nominal
anchor. $B_t/P_t$ ranges from 0.14994 to 0.76711; the slight undershoot of
0.15 is integer rounding. The five constrained arms share $B_t$ within task.

Managed-state cost tokenizes the joined rendered message-plan objects once,
including pointers, with the frozen tokenizer for
\texttt{Qwen/Qwen2.5-Coder-32B-Instruct}, revision
\texttt{381fc969f78e}\ldots. It is not the sum of independently tokenized
objects, nor the full provider request including all protocol overhead.
The API's recorded prompt usage is a separate outcome. Absolute budgets from
the earlier host accounting and the later Qwen accounting are not directly
comparable token units.

\subsection{Delivered-state matching and measured resources}
Let $D_{t,a}$ be the median recorded managed-state token count across steps
for task $t$, arm $a$. The delivery diagnostic requires
$\max(D_{t,a},D_{t,b})/\min(D_{t,a},D_{t,b})-1\leq0.10$ for every task.
No capped-arm pair passes on all eight tasks (Table~\ref{tab:delivery}). This
is a median managed-state
diagnostic, not an equality test on full API input or all-service compute.

\begin{table}[!ht]
\centering\small
\caption{All ten capped-arm delivery comparisons. The last column is the
largest relative gap across the eight tasks, expressed in percent.}
\label{tab:delivery}
\begin{tabular}{@{}lrr@{}}
\toprule
Pair & Tasks within 10\% & Maximum gap (\%) \\
\midrule
FIFO--LRU-D & 7/8 & 17.5 \\
FIFO--GA & 6/8 & 18.7 \\
FIFO--UC & 0/8 & 88.8 \\
FIFO--OA & 5/8 & 36.1 \\
LRU-D--GA & 5/8 & 30.2 \\
LRU-D--UC & 0/8 & 88.8 \\
LRU-D--OA & 6/8 & 28.9 \\
GA--UC & 0/8 & 83.0 \\
GA--OA & 6/8 & 27.7 \\
UC--OA & 1/8 & 81.7 \\
 
\bottomrule
\end{tabular}
\end{table}

\begin{table}[!ht]
\centering\small
\caption{Resources in the eight complete blocks only. Main calls count
recorded agent LLM steps; importance and summary calls are separate counters.
Input sums the main trajectories' recorded API prompt usage, excluding
auxiliary requests. Wall time is median whole-trajectory seconds; ceiling
counts runs reaching 25 steps, not successful completions.}
\label{tab:resources}
\begin{tabular}{@{}lrrrrrr@{}}
\toprule
Arm & Main calls & Importance & Summary & Input tokens & Wall (s) & Ceiling \\
\midrule
Full & 181 & 0 & 0 & 2,140,636 & 131.7 & 6/8 \\
FIFO & 191 & 0 & 0 & 1,147,832 & 103.0 & 7/8 \\
LRU-D & 181 & 0 & 0 & 1,074,649 & 99.5 & 4/8 \\
GA & 198 & 285 & 0 & 1,206,642 & 143.9 & 7/8 \\
UC & 197 & 0 & 0 & 1,091,350 & 102.0 & 7/8 \\
OA & 190 & 0 & 169 & 1,166,383 & 154.4 & 7/8 \\
 
\bottomrule
\end{tabular}
\end{table}

GA's paired whole-trajectory wall-time difference from FIFO is positive on
all eight tasks: mean $+67.45$ seconds, median $+43.47$ seconds. The difference
between the two arm medians is instead $40.89$ seconds. These are distinct
summaries. Whole-trajectory differences include changed execution paths,
tool work, and auxiliary work; they do not isolate importance-rating latency
or prove a serving-layer slowdown. CPU embedding work is not an LLM call.
OA's 169 summary calls and GA's 285 importance calls in
Table~\ref{tab:resources} prevent interpreting
same-cap outcomes as equal-compute comparisons.

\paragraph{Dollar estimates are unresolved, not billed measurements.}
The full-run \texttt{run\_result.json} stores an estimate of \$213.52,
whereas the associated \texttt{cost.json} ledger stores \$235.99. The ledger
uses preset prices of \$20/\$100 per million input/output tokens for the
recorded alias, explicitly conservative assumptions rather than provider
billing. The two totals have not been reconciled; neither is used as an
actual-spend result or combined with the eight-block table above.

\FloatBarrier
\section{Measurement Validity and Reproducibility}
\label{app:validity}
\setcounter{table}{0}

\subsection{Lifecycle signals are implementation observations}
In task \texttt{sympy\_\_sympy-19007}, a read of unchanged source content
creates a new disk artifact and advances its version. The file is
\texttt{sympy/matrices/expressions/blockmatrix.py}. The full-source sidecar
hashes in Table~\ref{tab:read_diagnostic} establish equal content across
each edit/read pair; different line-numbered file views are not compared as
if they were full-source snapshots.

\begin{table}[!ht]
\centering\small
\caption{Diagnostic case from the saved full-context trajectory. All four
rows are disk-only records, not counted context objects. Hash prefixes are
SHA-256 of complete saved source content.}
\label{tab:read_diagnostic}
\begin{tabular}{@{}rllrl@{}}
\toprule
Step & Object & Source action & Version & Content hash prefix \\
\midrule
3 & art\_0002 & edit\_file & 2 & \texttt{907377b09b27} \\
5 & art\_0003 & read\_file & 3 & \texttt{907377b09b27} \\
6 & art\_0005 & edit\_file & 4 & \texttt{e52604081c45} \\
14 & art\_0008 & read\_file & 5 & \texttt{e52604081c45} \\
 
\bottomrule
\end{tabular}
\end{table}

At step 14, the unchanged read produces five dependency-change invalidations
of tool-output objects. The tool's read path returns both a view and full
artifact content; the loop creates a versioned disk artifact before adding
the context view. Without a content-equality guard, a recorded version event
is not proof of a semantic change. This case invalidates that interpretation
of recorded stale flags; it does not estimate the defect's frequency over
the corpus. No corrected-system trajectories have been generated.

The access clock has a separate problem: the host updates
\texttt{last\_access\_step} and \texttt{access\_count} for
\texttt{INCLUDE\_IN\_PROMPT}, including pointer objects. The next policy
decision therefore sees recent inclusion rather than necessarily recent
demand. The later LRU-D selector uses a separate demand observer; inherited
OA retains the old counters. We do not infer a benefit of true recency or
reuse from this implementation. These limitations qualify the policy
comparisons as well as the lifecycle interpretation.

\subsection{Protocol history and omitted conclusions}
Table~\ref{tab:protocol} summarizes the protocol history.

\begin{table}[!ht]
\centering\small
\caption{Declarations and revisions are not all prospective preregistration.
Dates are the dates stated in the archived documents, not a new audit of
their entire commit history.}
\label{tab:protocol}
\begin{tabularx}{\linewidth}{@{}l Y@{}}
\toprule
Record & Status and implication \\
\midrule
Original, Aug.\ 14 & Declares calibration/held-out use, fixed policy
parameters, and nominal-budget repeated calls as the primary policy endpoint. \\
A10.4, Aug.\ 27 & Defines same-cap comparisons and treats delivered tokens as
an outcome; it does not establish equal delivered context. \\
A10.5, Aug.\ 27 & Introduces the two-endpoint framing after shakedown evidence.
This framing is development-informed. \\
A10.6, Aug.\ 27 & Adds explicit mechanism-exhaustion handling after a prior
calibration-run failure; explicitly not preregistered. \\
A10.7, Aug.\ 28 & Reactive serving-protocol amendment after a shakedown
failure; also not preregistered. It is not a new agent-quality replication. \\
\bottomrule
\end{tabularx}
\end{table}

Content-reference detection measures matching signatures, not causal utility;
its coverage differs by type and its annotator checks did not provide human
ground truth. Those observations and old ablation summaries are not used to
claim that inexpensive objects have greater future utility. Similarly, a
bootstrap interval and non-significant test do not establish policy equivalence.
Small conditional samples, dependent objects within tasks, and one follow-up
run per task--arm leave substantial uncertainty.

The later 48 complete task--arm cells lack valid formal repair evaluation.
An agent finish flag cannot fill this gap. Separate serving replays are also
excluded from performance claims: a purported cold-reset probe processes a
real request and changes subsequent cache/queue state. Removing its first
measurement does not restore an unaffected run. No cache benefit, serving
speedup, or corrected implementation result is claimed.

\subsection{Exploratory serving replay: platform and scope}
\label{app:hardware}
The serving-feasibility observation in Section~\ref{sec:resources} comes from
an exploratory replay on an NVIDIA GB10 host: Ubuntu 24.04.3, aarch64, CUDA
13.0, and 121.6\,GB usable unified memory, serving Qwen2.5-Coder-32B under a
frozen 32,768-token context limit. Execution was replay-only, re-serving saved
trajectory prompts; no new agent runs were performed. The overflow figures are
a static tokenization of those prompts computed before the serving engine
started, so they do not depend on the replay's timing behaviour.

This replay is exploratory and is not matched to the paper's main experimental
units: it covers a small set of calibration tasks rather than the held-out
split or the eight complete six-arm blocks, and it uses a different served
model from the agent runs. Its task-level serving measurements are therefore
not comparable to the policy contrasts reported in the main text.

No sustained system-memory-pressure regime was reached at the tested scale.
Recorded peak process resident memory stayed near 1.6\,GB with no swap use, and
a separate concurrency ladder recorded no sustained swap-out, with the machine
limited by scheduling and admission rather than memory. These runs therefore
support no physical-memory-efficiency claim for any policy, and none is made.
Per-policy latency, prefix-cache, and KV-utilisation figures from the same
replay are affected by the probe defect above and by unequal trajectory
lengths, and are not reported.

\subsection{What the archive supports}
\begin{table}[!ht]
\centering\small
\caption{Evidence locators within the working archive and their limits.
No anonymous public release is claimed. Prefix A denotes \texttt{attempts1/}; G denotes
\texttt{generative-agents-eval/}.}
\label{tab:repro_scope}
\begin{tabularx}{\linewidth}{@{}l Y@{}}
\toprule
Evidence & Locator and supported check \\
\midrule
Object accounting & A: \path{results/traces/} and
\path{results/tables/cost.csv}; saved-result aggregation only. \\
Earlier policy outcomes & A: cell files in \path{results/policy_tight/},
\path{results/policy_w3/}, and \path{results/policy_heldout/}.
Keep sweep identity when pairing. \\
Later outcomes & G: \path{results/phase3_calibration_v3/agent/};
derived cross-check: \path{results/phase3_quality_v1/QUALITY.json}. \\
Frozen mechanisms & G: \path{results/budget_freeze_v2/budget.json},
\path{src/policies/}, \path{src/prompts/}, and tokenizer/embedding constants. \\
Recalculation & Manuscript: \path{analysis/build_appendix_evidence.py};
standard-library-only, input hashes, generated table rows; no model calls. \\
\bottomrule
\end{tabularx}
\end{table}

The archived analysis dependency specification lists Python 3.12.7, NumPy
1.26.4, SciPy 1.13.1, and pandas 2.2.2 for the earlier statistics; the follow-up
has a separate requirements file and pinned local tokenizer/embedding revisions.
These specifications are not evidence that every historical runtime was
identical. Our exact enumeration avoids silently changing the treatment of
zero differences with SciPy versions. Full provider request/retry provenance,
the served model revision, and outside intervention history are incomplete.
Within-run caches stabilize some repeated auxiliary work but do not make
cloud generations exactly reproducible across reruns. The supplied audit
script, derived tables, and input hashes support the saved-result checks
stated in Table~\ref{tab:repro_scope}; we promise neither a complete anonymous
raw-data release nor byte-identical trajectory regeneration.

\clearpage
\section*{NeurIPS Paper Checklist}
\label{checklist:start}

\begin{enumerate}

\begin{samepage}
\item {\bf Claims}
    \item[] Question: Do the main claims made in the abstract and introduction accurately reflect the paper's contributions and scope?
    \item[] Answer: \answerYes{}
\end{samepage}
    \item[] Justification: The abstract and Section~\ref{sec:intro} state three scoped contributions: (1) a characterization of semantic heterogeneity in coding-agent working memory through typed-object accounting of size, retention, representation, and compression behavior; (2) two semantically informed management strategies examined as case studies; and (3) a four-level evaluation framework separating stored state, delivered context, management work, and task/process outcome. Sections~\ref{sec:characterization}--\ref{sec:lessons} distinguish descriptive memory structure from semantic utility and process metrics from repair success, and do not claim general policy superiority.

\begin{samepage}
\item {\bf Limitations}
    \item[] Question: Does the paper discuss the limitations of the work performed by the authors?
    \item[] Answer: \answerYes{}
\end{samepage}
    \item[] Justification: Sections~\ref{sec:setup}--\ref{sec:lessons} and Appendices~\ref{app:setup}--\ref{app:validity} discuss conditional samples, limited repetitions, budget and delivery differences, lifecycle-signal defects, missing valid repair evaluation, and incomplete model/request provenance. These limits qualify the policy results as well as their generalization.

\begin{samepage}
\item {\bf Theory assumptions and proofs}
    \item[] Question: For each theoretical result, does the paper provide the full set of assumptions and a complete (and correct) proof?
    \item[] Answer: \answerNA{}
\end{samepage}
    \item[] Justification: The paper contains no theoretical results or claimed optimality or feasibility guarantees. Its equations define accounting quantities and implemented policy scores; the recorded greedy floor rule is explicitly not a proof of global infeasibility (Appendices~\ref{app:retrieval} and~\ref{app:resources}).

\begin{samepage}
    \item {\bf Experimental result reproducibility}
    \item[] Question: Does the paper fully disclose all the information needed to reproduce the main experimental results of the paper to the extent that it affects the main claims and/or conclusions of the paper (regardless of whether the code and data are provided or not)?
    \item[] Answer: \answerNo{}
\end{samepage}
    \item[] Justification: Appendices~\ref{app:calibration}, \ref{app:retrieval}, and~\ref{app:validity} document saved-record aggregation and statistical recomputation, but the served model revision and complete historical requests, retries, and intervention records are unavailable. The stated saved-result checks therefore do not constitute full experimental reproducibility.

\begin{samepage}
\item {\bf Open access to data and code}
    \item[] Question: Does the paper provide open access to the data and code, with sufficient instructions to faithfully reproduce the main experimental results, as described in supplemental material?
    \item[] Answer: \answerNo{}
\end{samepage}
    \item[] Justification: Appendix~\ref{app:validity} locates evidence in the working archive, but the manuscript does not provide a complete anonymous public code/data release or a standalone raw-trace package. The included audit script, input hashes, and derived tables are not substitutes for access to the experimental inputs.

\begin{samepage}
\item {\bf Experimental setting/details}
    \item[] Question: Does the paper specify all the training and test details (e.g., data splits, hyperparameters, how they were chosen, type of optimizer) necessary to understand the results?
    \item[] Answer: \answerNo{}
\end{samepage}
    \item[] Justification: Sections~\ref{sec:setup}, \ref{sec:policies}, and~\ref{sec:resources} and Appendices~\ref{app:setup}--\ref{app:resources} report the task splits, selection rules, recorded policy parameters, tools, budgets, and stopping rules, but a complete historical parameter-selection ledger and serving/decoding provenance are not available. There is no model training; the missing details prevent an unqualified claim that all experimental settings are specified.

\begin{samepage}
\item {\bf Experiment statistical significance}
    \item[] Question: Does the paper report error bars suitably and correctly defined or other appropriate information about the statistical significance of the experiments?
    \item[] Answer: \answerYes{}
\end{samepage}
    \item[] Justification: Appendices~\ref{app:calibration} and~\ref{app:retrieval} report task-paired effects, exact signed-rank tests, Holm correction, and explicitly post-hoc 95\% task-bootstrap intervals, with the resampling method and seed specified. These intervals describe task-level variation, not repeated-model-run variability; the object characterization is descriptive rather than a population-wide significance claim.

\begin{samepage}
\item {\bf Experiments compute resources}
    \item[] Question: For each experiment, does the paper provide sufficient information on the computer resources (type of compute workers, memory, time of execution) needed to reproduce the experiments?
    \item[] Answer: \answerNo{}
\end{samepage}
    \item[] Justification: Appendix~\ref{app:resources} separates recorded main/auxiliary calls, input tokens, wall time, and incomplete-run accounting, and discloses conflicting dollar estimates. Appendix~\ref{app:hardware} gives the NVIDIA GB10 hardware and memory environment for the exploratory serving replay. A complete per-experiment hardware/memory specification and total compute inventory for the main archived agent runs, including provider-side resources and development runs, remain unavailable.
    
\begin{samepage}
\item {\bf Code of ethics}
    \item[] Question: Does the research conducted in the paper conform, in every respect, with the NeurIPS Code of Ethics \url{https://neurips.cc/public/EthicsGuidelines}?
    \item[] Answer: \answerYes{}
\end{samepage}
    \item[] Justification: The authors confirm that they have reviewed the NeurIPS Code of Ethics and that the research reported in this paper conforms to it. Sections~\ref{sec:resources}--\ref{sec:lessons} and Appendix~\ref{app:validity} disclose the measurement and reproducibility limitations relevant to research transparency.

\begin{samepage}
\item {\bf Broader impacts}
    \item[] Question: Does the paper discuss both potential positive societal impacts and negative societal impacts of the work performed?
    \item[] Answer: \answerNo{}
\end{samepage}
    \item[] Justification: The manuscript discusses measurement validity, resource accounting, and evaluation limitations, but does not provide a discussion covering both positive and negative societal impacts. Its narrow empirical scope is not a claim that downstream coding-agent deployment has no societal impact.
    
\begin{samepage}
\item {\bf Safeguards}
    \item[] Question: Does the paper describe safeguards that have been put in place for responsible release of data or models that have a high risk for misuse (e.g., pre-trained language models, image generators, or scraped datasets)?
    \item[] Answer: \answerNA{}
\end{samepage}
    \item[] Justification: This manuscript does not release a new pretrained model or a high-risk dataset requiring controlled access. The reported work analyzes saved coding-agent trajectories and does not claim that safeguards for a deployed agent have been evaluated.

\begin{samepage}
\item {\bf Licenses for existing assets}
    \item[] Question: Are the creators or original owners of assets (e.g., code, data, models), used in the paper, properly credited and are the license and terms of use explicitly mentioned and properly respected?
    \item[] Answer: \answerNo{}
\end{samepage}
    \item[] Justification: The manuscript credits the benchmark and prior methods and identifies relevant model/tokenizer assets, but does not enumerate all asset licenses, dataset/repository terms, and provider terms of use. Complete license and terms-of-use compliance has not been verified in this manuscript.

\begin{samepage}
\item {\bf New assets}
    \item[] Question: Are new assets introduced in the paper well documented and is the documentation provided alongside the assets?
    \item[] Answer: \answerNo{}
\end{samepage}
    \item[] Justification: The audit script and derived tables are accompanied by input and limitation documentation, but a complete license and release/usage documentation for the new research artifacts is not included. No public benchmark or trained-model release is claimed (Appendix~\ref{app:validity}).

\begin{samepage}
\item {\bf Crowdsourcing and research with human subjects}
    \item[] Question: For crowdsourcing experiments and research with human subjects, does the paper include the full text of instructions given to participants and screenshots, if applicable, as well as details about compensation (if any)? 
    \item[] Answer: \answerNA{}
\end{samepage}
    \item[] Justification: The reported analyses use existing benchmark tasks and saved software-agent traces; no recruited participants, crowdsourcing experiment, or human-subject study is reported. Consequently, participant instructions, screenshots, and compensation do not apply to these reported experiments.

\begin{samepage}
\item {\bf Institutional review board (IRB) approvals or equivalent for research with human subjects}
    \item[] Question: Does the paper describe potential risks incurred by study participants, whether such risks were disclosed to the subjects, and whether Institutional Review Board (IRB) approvals (or an equivalent approval/review based on the requirements of your country or institution) were obtained?
    \item[] Answer: \answerNA{}
\end{samepage}
    \item[] Justification: The reported work does not include a crowdsourcing or human-subject experiment for which participant-risk disclosures or an IRB approval are being claimed. This answer is limited to the study described in the manuscript, not a certification of unrelated historical activities.

\begin{samepage}
\item {\bf Declaration of LLM usage}
    \item[] Question: Does the paper describe the usage of LLMs if it is an important, original, or non-standard component of the core methods in this research? Note that if the LLM is used only for writing, editing, or formatting purposes and does \emph{not} impact the core methodology, scientific rigor, or originality of the research, declaration is not required.
    \item[] Answer: \answerYes{}
\end{samepage}
    \item[] Justification: Sections~\ref{sec:setup}, \ref{sec:policies}, and~\ref{sec:resources} and Appendices~\ref{app:setup}, \ref{app:calibration}, \ref{app:retrieval}, and~\ref{app:resources} describe the coding-agent LLM, summary generation, and importance scoring, including the recorded model alias and auxiliary-call accounting. Appendix~\ref{app:validity} explicitly identifies the incomplete historical model/request provenance.

\end{enumerate}

\end{document}